\documentclass{edm_article}

\usepackage{booktabs}

\begin{document}

\title{ALL-IN-ONE: Multi-Task Learning BERT models for Evaluating Peer Assessments}
%\subtitle{[Extended Abstract]
%\titlenote{A full version of this paper is available as
%\textit{Author's Guide to Preparing ACM SIG Proceedings Using
%\LaTeX$2_\epsilon$\ and BibTeX} at
% Submissions for EDM are double-blind: please do not include any
% author names or affiliations in the submission. 
% Anonymous authors:
\numberofauthors{1}
\author{
Qinjin Jia, Jialin Cui, Yunkai Xiao, Chengyuan Liu, Parvez Rashid, Edward Gehringer\\
       \affaddr{Department of Computer Science} \\
       \affaddr{North Carolina State University} \\
       \affaddr{Raleigh, NC, USA} \\
       \email{qjia3, jcui9, yxiao28, cliu32, mrashid4, efg@ncsu.edu}
}

\maketitle

%\onecolumn

\begin{abstract}
Peer assessment has been widely applied across diverse academic fields over the last few decades, and has demonstrated its effectiveness. However, the advantages of peer assessment can only be achieved with high-quality peer reviews. Previous studies have found that high-quality review comments usually comprise several features (e.g., contain suggestions, mention problems, use a positive tone). Thus, researchers have attempted to evaluate peer-review comments by detecting different features using various machine learning and deep learning models. However, there is no single study that investigates using a multi-task learning (MTL) model to detect multiple features simultaneously. This paper presents two MTL models for evaluating peer-review comments by leveraging the state-of-the-art pre-trained language representation models BERT and DistilBERT. Our results demonstrate that BERT-based models significantly outperform previous GloVe-based methods by around 6\% in F1-score on tasks of detecting a single feature, and MTL further improves performance while reducing model size.
\end{abstract}

%% A category with the (minimum) three required fields
%%A category including the fourth, optional field follows...
%\category{D.2.8}{Software Engineering}{Metrics}[complexity measures, performance measures]


\section{Introduction}

Peer assessment is a process by which students give feedback on other students' work based on a rubric provided by the instructor \cite{sadler2006impact, topping1998peer}. This assessment strategy has been widely applied across diverse academic fields, such as computer science \cite{wang2012assessment}, medicine \cite{violato2006self}, and business \cite{brutus2013can}. Furthermore, massive open online courses (MOOCs) commonly use peer assessment to provide feedback to students and assign grades. There is abundant literature \cite{double2020impact, topping1998peer, van2010effective, li2020does} demonstrating the efficacy of peer assessment. For example, Doubling et al. \cite{double2020impact} conducted a meta-analysis of 54 controlled experiments for evaluating the effect of peer assessment across subjects and domains. The results indicate that peer assessment is more effective than teacher assessment, and also remarkably robust across a wide range of contexts \cite{double2020impact}.

However, low-quality peer reviews are a persistent problem in peer assessment, and considerably weaken the learning effect \cite{omar2018use, suen2014peer}. The advantages of peer assessment can only be achieved with high-quality peer reviews \cite{nelson2009nature}. This suggests that peer reviews should not be simply transmitted to other students but rather should be vetted in some way. Course staff could check the quality of each review comment\footnote{In some peer-assessment systems, reviews are ``holistic".  In others, including the systems we are studying, each review contains a set of review comments, each comment gives a response to a different criterion in the rubric.} and assess its credibility manually, but this is not efficient. Sometimes (e.g., for MOOCs), this is not remotely possible. Therefore, to ensure the quality of peer reviews and the efficiency of evaluating their quality, the peer-assessment platform should be capable of assessing peer reviews automatically. We call this Automated Peer-Review Evaluation.

Previous research has determined that high-quality review comments usually comprise several features \cite{nelson2009nature, caligiuri2013editors, van2010effective}. Examples of such features are, ``contains suggestions'', ``mentions problems'', ``uses a positive tone'', ``is helpful'', ``is localized'' \cite{nelson2009nature}. Thus, one feasible and promising way to evaluate peer reviews automatically is to adjudicate the quality of each review comment based on whether it comprises the predetermined features, by treating this task as a text classification problem. If a peer-review comment does not contain some of the features, the peer-assessment platform could suggest that the reviewer should revise the review comment to add missing features.  Additionally, containing suggestions, mentioning problems, and using a positive tone, are among the most essential features. Thus, we use them for this study.

\begin{figure*}
\centering
\includegraphics[width=0.90\textwidth]{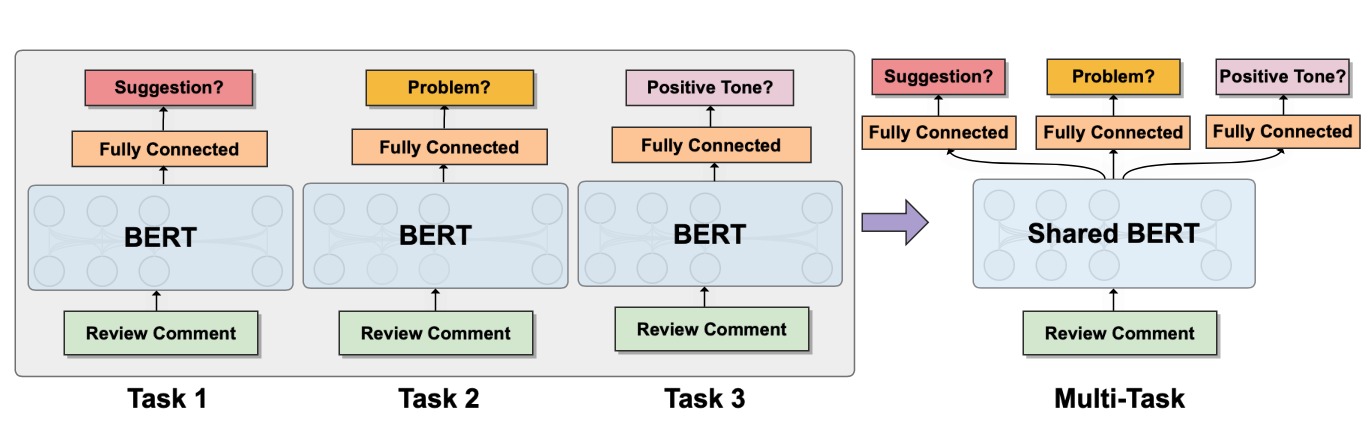}
\caption{Illustration of the single-task and multi-task learning settings}
\label{fig:illustration}
\end{figure*}

Previous work for automatically evaluating review comments has focused on tasks that detect a single feature. For example, Xiong and Litman \cite{xiong2011automatically} designed sophisticated features and used traditional machine-learning methods for identifying peer-review helpfulness. Zingle et al. \cite{zingle2019detecting} utilized different rule-based, machine-learning, and deep-learning methods for detecting suggestions in peer-review comments. However, to the best of our knowledge, no single study exists that investigates using a multi-task learning (MTL) model to detect multiple features simultaneously (as illustrated in Figure \ref{fig:illustration}), albeit extensive research has been carried out on the topic of automated peer-review evaluation (e.g., \cite{xiong2010assessing, xiong2010identifying, xiong2011automatically, xiao2020detecting, xiaoproblem, zingle2019detecting, negi2017suggestion, ramachandran2013automated, fromm2020argument}).

There are at least two motivations for using multi-task learning (MTL) to detect features simultaneously. Firstly, the problem naturally leads itself well to MTL, due to multiple features usually needing to be employed for a comprehensive and precise evaluation of peer-review comments. If we treat this MTL problem as multiple independent single tasks, total model size and prediction time will increase by a factor of the number of features used for evaluating review comments. Secondly, MTL can increase data efficiency. This implies that learning tasks jointly can lead to performance improvement compared with learning them individually, especially when training samples are limited \cite{crawshaw2020multi, zhang2021survey}. More specifically, MTL can be viewed as a form of inductive transfer learning, which can help improve the performance of each jointly learned task by introducing an inductive bias \cite{caruana1997multitask}.

Additionally, the pre-trained language model, BERT (Bidirectional Encoder Representations from Transformers) \cite{devlin2018bert}, has become a standard tool for reaching the state of the art in many natural language processing (NLP) tasks. BERT can significantly reduce the need for labeled data. Therefore, we propose multi-task learning (MTL) models for evaluating review comments by leveraging the state-of-the-art pre-trained language representation models BERT and DistilBERT. We first compare a BERT-based single-task learning (STL) model with the previous GloVe-based STL model. We then propose BERT and DistilBERT based MTL models for jointly learning different tasks simultaneously. 

The rest of the paper is organized as follows: Section 2 presents related work. Section 3 describes the dataset used for this study. The proposed single-task and multi-task text classification models are elaborated in Section 4. Section 5 details the experimental setting and results. In Section 6, we conclude the paper, mention the limitations of our research, and discuss future work.

\section{Related Work}

\subsection{Automated Peer-Review Evaluation}

The earliest study on automated peer-review evaluation was performed by Cho in 2008 \cite{cho2008machine}. They manually broke down every peer review comment into review units (self-contained messages in each review comment) and then coded them as praise, criticism, problem detection, solution suggestion. Cho \cite{cho2008machine} utilized traditional machine learning methods, including naive Bayes, support vector machines (SVM), and decision trees, to classify the review units.

Xiong et al. attempted to use features (e.g., counts of nouns, verbs) derived from regular expressions and dependency parse trees and rule-based methods to detect localization in the review units \cite{xiong2010identifying}. Then, they designed more sophisticated features by combining generic linguistic features mined from review comments and specialized features, and used SVM to identify peer-review helpfulness \cite{xiong2011automatically}. After that, Xiong et al. upgraded their models to comment-level (use whole review comment instead of review units as the input) \cite{nguyen2014classroom, nguyen2016instant}.   

Then, researchers started to use deep neural networks on tasks of automated peer-review evaluation for improving accuracy. Zingle et al. compared rule-based machine-learning and deep neural-network methods for detecting suggestions in peer assessments, and the result showed that deep-learning methods outperformed other traditional methods \cite{zingle2019detecting}. Xiao et al. collected around 20,000 peer-review comments and leveraged different neural networks to detect problems in peer assessments \cite{xiao2020detecting}.

\vspace{-1pt}

\subsection{Multi-Task Learning}

Multi-task learning (MTL) is an important subfield of machine learning in which multiple tasks are learned simultaneously \cite{zhang2017survey, crawshaw2020multi, caruana1997multitask} to help improve the generalization performance of all the tasks. A task is defined as $\{p(x), p(y|x), L)\}$, where $p(x)$ is the input distribution, $p(y|x)$ is the distribution over the labels given the inputs, and $L$ is the loss function. For the MTL setting in this paper, all tasks have the same input distribution $p(x)$ and loss function $L$, but different distributions over the labels given the inputs $p(y|x)$.

In the context of deep learning, all methods of MTL can be partitioned into two groups: hard-parameter sharing and soft-parameter sharing \cite{caruana1997multitask}. For hard-parameter sharing, the hidden layers are shared between all tasks while keeping several task-specific output layers. For soft-parameter sharing, each task has its independent model, but the distance between the different models' parameters is regularized. For this study, we use the hard-parameter sharing approach.

\begin{table*}
\centering
\caption{Sample Rubric Criteria}
\begin{tabular}{ll}
\hline
Does the design incorporate all of the functionality required?                                       \\
Have the authors converted all the cases discussed in the test plan into automated tests?                \\
Does the design appear to be sound, following appropriate principles and using appropriate patterns? \\
\hline
\end{tabular}
\label{table:samplecriteria}
\end{table*}

\begin{table*}
\centering
\caption{Sample Data}
\begin{tabular}{llll}
\hline
\textbf{Peer-Review Comments} (lower-cased) & Sugg. & Prob. & Tone \\ \hline\hline
\small{lots of good background details is given but the testing and implementation sections are missing.}               & 0    &  1    &  1   \\
\small{the explanation is clear to follow but it could also include some explanation of the use cases.}                      &  1    &  0   &  1   \\
\small{only problem statement is explained and nothing about design. please add design and diagrams.}                       &  1    &   1   &  0  \\ \hline
\end{tabular}
\label{table:sampledata}
\end{table*}

\section{Data}

\subsection{Data Source: Expertiza}
The data in this study is collected from an NSF-funded peer-assessment platform, Expertiza\footnote{https://github.com/expertiza/expertiza}. In this flexible peer-assessment system, students can submit their work and peer-review the learning objects (such as articles, code, and websites) of other students \cite{gehringer2007reusable}. This platform supports multi-round peer review. In the assignments that provided the review comments for this study, two rounds of peer review (and one round of meta-review) were used:

\vspace{-10pt}

\begin{enumerate}
    \item The \textit{formative-feedback} phase: For the first round of review, students upload substantially complete projects. The system then assigns each student to review a set number of these submissions, based on a rubric provided. Sample rubric criteria are provided in Table \ref{table:samplecriteria}.
    
    \item The \textit{summative-feedback} phase: After students have had an opportunity to revise their work based on feedback from their peers, final deliverables are submitted and peer-reviewed using a summative rubric. The rubric may include criteria such as ``How well has the team has addressed the feedback given in the first review round?''. Many criteria in the rubric ask reviewers to provide a numeric rating as well as a textual comment.
    
    \item The \textit{meta-review} phase: After the grading period is over, course staff typically assess and grade the reviews provided by students.
\end{enumerate}

\vspace{-10pt}

For this study, all textual responses to the rubric criteria from the formative-feedback phase and the summative-feedback phase of a graduate-level software-engineering course are extracted to constitute the dataset. Each response to a rubric criterion constitutes a peer-review comment. All responses from one student to a set of criteria in a single rubric are called a peer review or a review. In this study, we focus on evaluating each peer-review comment. After filtering out review comments that only contain symbols and special characters, the dataset consists of 12,053 review comments. In the future, we will update the platform, and this type of review comments will be rejected by the system directly.

\subsection{Annotation Process}

One annotator who is a fluent English speaker and familiar with the course context annotated the dataset. For quality control, 100 reviews were randomly sampled from the dataset and labeled by a second annotator who is also a fluent English speaker and familiar with the course context. The inter-annotator agreement between two annotators was measured by Cohen's $\kappa$ coefficient, which is generally thought to be a more robust measure than simple percent agreement calculation \cite{mchugh2012interrater}. Cohen's $\kappa$ coefficient for each label is shown in Table \ref{table:cohen}. The result suggests that the two annotators had almost perfect agreement (>0.81) \cite{mchugh2012interrater}. Sample annotated comments are provided in Table \ref{table:sampledata}.

\begin{table}[]
\caption{Inter-Annotator Agreement (Cohen's $\kappa$)}
\begin{tabular}{l|lll|l}
\hline
Label         & Suggestion & Problem & Tone & Average \\ \hline
Cohen's Kappa & 0.92       & 0.84    & 0.87 & 0.88    \\ \hline
\end{tabular}
\label{table:cohen}
\vspace{-8pt}
\end{table}

We define each feature (label) in the context of automated peer-review evaluation as follows:

\vspace{-5pt}

\textbf{Suggestion:} A comment is said to contain a suggestion if it mentions how to correct a problem or make improvements.

\vspace{-5pt}

\textbf{Problem:} A comment is said to detect problems if it points out something that is going wrong in peers' work.

\vspace{-5pt}

\textbf{Positive Tone:} A comment is said to use a positive tone if it has an overall positive semantic orientation.

\subsection{Statistics on the Dataset}
The minority class for each label includes more than 20\% of samples, and thus the dataset is mildly imbalanced. It consists of 12,053 peer-review comments, and the average number of words for each peer-review comment is 29. We found that most students (over three-quarters) use a positive tone in their peer-review comments. Around half of the review comments mention problems with their peers' work, but only one-fifth of review comments give suggestions. Characteristics of the dataset are shown in Table \ref{table:dataset} below,

\begin{table}[h]
\caption{Statistics on the Dataset}
\vspace{1.5pt}
\begin{tabular}{lllll}
\toprule
Label     & Class & \%samples & avg.\#words & max\#words \\ \midrule
Sugg.     & 0     & 79.2\%    & 22           & 922        \\
          & 1     & 20.8\%    & 58           & 1076       \\ \midrule
Prob.     & 0     & 56.7\%    & 22           & 479        \\
          & 1     & 43.3\%    & 38           & 1076       \\ \midrule
Pos. Tone & 0     & 22.2\%    & 28           & 1040       \\
          & 1     & 77.8\%    & 29           & 1076       \\
\bottomrule  
\end{tabular}
\label{table:dataset}
\end{table}

\section{Methodology}
In this section, we first briefly introduce Transformer \cite{vaswani2017attention}, BERT \cite{devlin2018bert}, and DistilBERT \cite{sanh2019distilbert}. Then we describe BERT and DistilBERT based single-task and multi-task models.

\vspace{-5pt}

\subsection{Transformer}
\begin{figure}[h]
\centering
\includegraphics[width=0.35\textwidth]{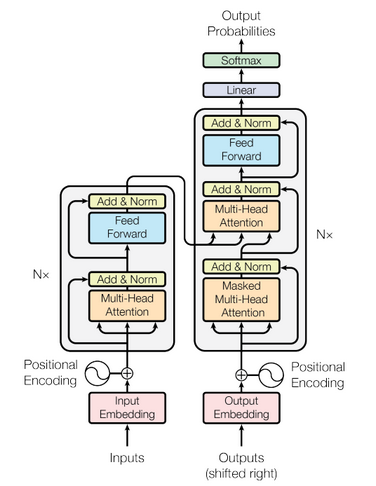}
\caption{Architecture of the Transformer \cite{vaswani2017attention}}
\label{fig:transformer}
\end{figure}

In 2017, Vaswani et al. published a groundbreaking paper, ``Attention is all you need," and proposed an architecture called Transformer, which significantly improved the performance of sequence-to-sequence tasks (e.g., machine translation) \cite{vaswani2017attention}. The Transformer is entirely built upon self-attention mechanisms without using any recurrent or convolutional layers. As shown in Figure \ref{fig:transformer}, the Transformer consists of two parts: the left part is an \textit{encoder}, and the right part is a \textit{decoder}. The \textit{encoder} block takes a batch of sentences represented as sequences of word IDs. Then the sequences pass through an embedding layer, and the positional embedding adds positional information of each word.

The \textit{encoder} block is then briefly introduced since BERT reuses it. Each \textit{encoder} consists of two layers: a multi-head attention layer and a feed-forward layer. The multi-head attention layer uses the self-attention mechanism, which encodes each word's relationship with every other word in the same sequence, paying more attention to the most relevant ones. For example, the output of this layer for the word ``like" in the sentence, ``we like the Educational Data Mining conference 2021!" will depend on all the words in the sentence. However, it will probably pay more attention to the word ``we" than to the words ``data" or ``mining."

\subsection{BERT}
BERT is a state-of-the-art pre-trained language representation model proposed by Devlin et al. \cite{devlin2018bert}. It has advanced the state-of-the-art results in many NLP tasks and significantly reduced the need for labeled data by pre-training on unlabeled data over different pre-training tasks. Each BERT model consists of 12 \textit{encoder} blocks of the Transformer model. The input representation is constructed by summing the corresponding token and positional embeddings. The length of the output sequence is the same as the input length, and each input token has a corresponding representation in the output. The output of the first token `[CLS]' (a special token added to the sequence) is utilized as the aggregate representation of the input sequence for classification tasks \cite{devlin2018bert}.

The BERT framework consists of two steps: pre-training and fine-tuning. During pre-training, the model is trained on unlabeled data, BooksCorpus (800M words) and English Wikipedia (2,500M words), over two pre-training tasks, Masked language model (MLM) and Next sentence prediction (NSP). For fine-tuning, the BERT model is first initialized with the pre-trained parameters, and then all of the parameters are fine-tuned using labeled data from the downstream tasks (e.g., text classification). For this study, we use HuggingFace pre-trained BERT\footnote{https://huggingface.co/bert-base-uncased} to initialize models and then fine-tune models with annotated peer-review comments for automated peer-review evaluation tasks.

\subsection{DistilBERT}
Although BERT has shown remarkable improvements across various NLP tasks and can be easily fine-tuned for downstream tasks, one main drawback of BERT is that it is very compute-intensive (i.e., it takes a huge amount of parameters, $\sim$110M parameters). Therefore, researchers are attempting to apply different methods for compressing BERT, including pruning, quantization, and knowledge distillation \cite{gupta2020compression}. One of the compressed BERT models is called DistilBERT \cite{sanh2019distilbert}. DistilBERT is compressed from BERT by leveraging the knowledge distillation technique during the pre-training phase. The authors \cite{sanh2019distilbert} demonstrated that DistilBERT has 40\% fewer parameters and is 60\% faster than the original BERT while retaining 97\% of its language-understanding capabilities. We will investigate whether we can reduce model size while retaining performance for our task with DisilBERT.\footnote{https://huggingface.co/distilbert-base-uncased}

\begin{figure*}
\centering
\includegraphics[width=0.8\textwidth]{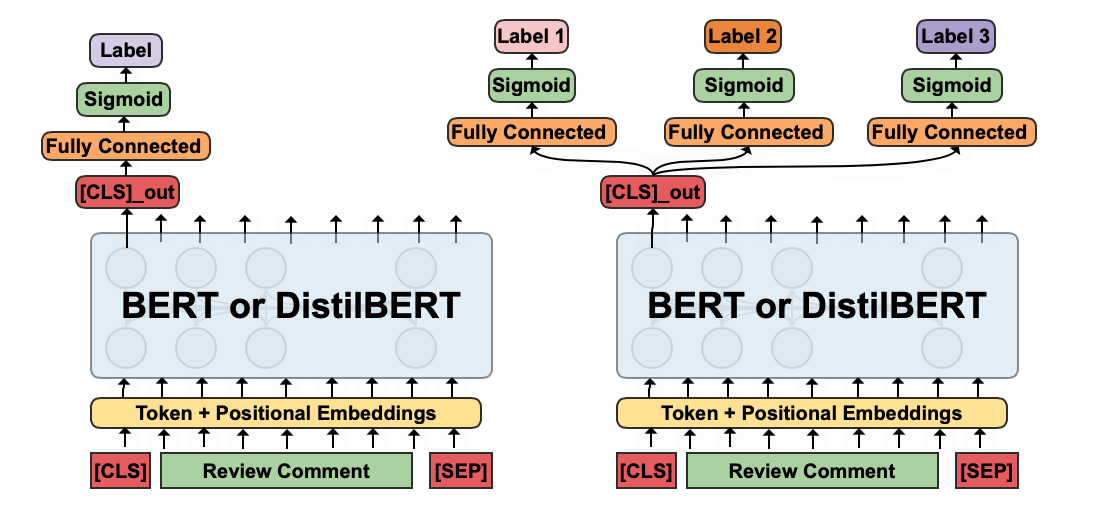}
\caption{BERT and DistilBERT based single-task and multi-task learning architectures}
\end{figure*}

\subsection{Input Preparation}
\textbf{Text Preprocessing:} First, URL links in peer-review comments are removed. Then, we lowercase all comments and leverage a spellchecker API\footnote{https://pypi.org/project/pyspellchecker/} to correct typos and misspellings. Finally, two special tokens ([CLS], [SEP]) are added to each review comment, as required for BERT. The [CLS] token is added to the beginning of each review for classification tasks. The [SEP] token is added at the end of each review.

\vspace{-8pt}

\textbf{Subword Tokenization:} The tokenizer used for BERT is a subword tokenizer called ``WordPiece'' \cite{wu2016google}. Traditional word tokenizers suffer the out-of-vocabulary (OOV) word problem. However, a subword tokenizer could alleviate the OOV problem. It splits a text into subwords, which then are converted to token IDs.

\vspace{-8pt}

\textbf{Input Representation:} The token IDs are padded or truncated to 100 for each sequence and then pass through a trainable embedding layer to be converted to token embeddings. The input representation for BERT is constructed by summing the token embeddings and positional embeddings.

\subsection{Single-Task and Multi-Task Models}

As mentioned in the BERT paper \cite{devlin2018bert} and other studies \cite{sun2019fine}, the pre-trained BERT model can be fine-tuned with just one additional output layer to create state-of-the-art models for a wide range of tasks, including text classification. Therefore, only one dense layer is added on top of the original BERT or DistilBERT model and used as a binary classifier for the single-task learning models. Three dense layers are added to the multi-task learning models, one for each label.

\begin{table*}[h]
\centering
\caption{Performance evaluation (average performance of 5 independent runs)}
\vspace{2pt}
\begin{tabular}{clclclclclclclclclclc}
\toprule
                                    & \multicolumn{3}{c}{\textbf{Suggestion}}                              & \multicolumn{3}{c}{\textbf{Problem}}                                 & \multicolumn{3}{c}{\textbf{Pos. Tone}}          \\
\cmidrule(r{4pt}){2-4} \cmidrule(l){5-7} \cmidrule(l){8-10}
 & Acc.            & Macro-F1      & \multicolumn{1}{l|}{AUC}           & Acc.            & Macro-F1      & \multicolumn{1}{l|}{AUC}           & Acc.            & Macro-F1      & AUC           \\
\midrule
\multicolumn{10}{l}{\textit{Training with 1000 labeled data samples}}                                                                                                                                                               \\
\multicolumn{1}{l|}{1 STL-GloVe (Baseline)}      & 82.0\%          & .744          & \multicolumn{1}{l|}{.865}          & 80.2\%          & .790          & \multicolumn{1}{l|}{.879}          & 76.0\%          & .700          & .823          \\
\multicolumn{1}{l|}{2 STL-BERT}       & 90.0\%          & .868          & \multicolumn{1}{l|}{\textbf{.975}} & \textbf{89.2\%} & \textbf{.892} & \multicolumn{1}{l|}{\textbf{.955}} & 87.0\%          & .828          & .940          \\
\multicolumn{1}{l|}{3 MTL-BERT}       & \textbf{94.0\%} & .904          & \multicolumn{1}{l|}{.974}          & 89.0\%          & .890          & \multicolumn{1}{l|}{\textbf{.955}} & \textbf{89.4\%} & \textbf{.846} & \textbf{.941} \\
\multicolumn{1}{l|}{4 STL-DistilBERT} & 92.4\%          & .890          & \multicolumn{1}{l|}{.970}          & 88.0\%          & .880          & \multicolumn{1}{l|}{.950} & 86.2\%          & .822          & .933          \\
\multicolumn{1}{l|}{5 MTL-DistilBERT} & 93.8\%          & \textbf{.910} & \multicolumn{1}{l|}{.971}          & 89.0\%          & .886          & \multicolumn{1}{l|}{.951}          & 88.6\%          & .824          & .939          \\
\midrule

\multicolumn{10}{l}{\textit{Training with 3000 labeled data samples}}                                                                                                                                                                        \\
\multicolumn{1}{l|}{1 STL-GloVe (Baseline)}      & 88.4\%          & .836          & \multicolumn{1}{l|}{.929}          & 83.0\%          & .830          & \multicolumn{1}{l|}{.898}          & 82.4\%          & .770          & .872           \\
\multicolumn{1}{l|}{2 STL-BERT}       & 93.8\%          & .910          & \multicolumn{1}{l|}{.980}          & 90.6\%          & .904          & \multicolumn{1}{l|}{\textbf{.964}} & 89.6\%          & .858          & \textbf{.948} \\
\multicolumn{1}{l|}{3 MTL-BERT}       & \textbf{94.6\%} & \textbf{.916} & \multicolumn{1}{l|}{\textbf{.981}} & \textbf{91.0\%} & \textbf{.906} & \multicolumn{1}{l|}{\textbf{.964}} & 90.0\%          & \textbf{.854} & .947          \\
\multicolumn{1}{l|}{4 STL-DistilBERT} & 94.0\%          & .910          & \multicolumn{1}{l|}{.979}          & 89.8\%          & .900          & \multicolumn{1}{l|}{.962}          & 89.0\%          & .850          & .942          \\
\multicolumn{1}{l|}{5 MTL-DistilBERT} & 94.2\%          & \textbf{.916} & \multicolumn{1}{l|}{.978}          & 89.6\%          & .892          & \multicolumn{1}{l|}{.960}          & \textbf{90.2\%} & .850          & .945          \\
\midrule

\multicolumn{10}{l}{\textit{Training with 5000 labeled data samples}}                                                                                                                                                                        \\
\multicolumn{1}{l|}{1 STL-GloVe (Baseline)}      & 89.9\%          & .852          & \multicolumn{1}{l|}{.947}          & 84.2\%          & .832          & \multicolumn{1}{l|}{.908}          & 85.0\%          & .794          & .883          \\
\multicolumn{1}{l|}{2 STL-BERT}       & 94.4\%          & .916          & \multicolumn{1}{l|}{.980}          & \textbf{91.2\%} & \textbf{.912} & \multicolumn{1}{l|}{\textbf{.968}} & 89.4\%          & .852          & .950          \\
\multicolumn{1}{l|}{3 MTL-BERT}       & \textbf{94.8\%} & \textbf{.922} & \multicolumn{1}{l|}{\textbf{.982}} & 91.0\%          & .908          & \multicolumn{1}{l|}{.966}          & \textbf{90.8\%} & \textbf{.854} & \textbf{.951} \\
\multicolumn{1}{l|}{4 STL-DistilBERT} & 94.2\%          & .912          & \multicolumn{1}{l|}{.978}          & 90.4\%          & .902          & \multicolumn{1}{l|}{.964}          & 89.8\%          & .860          & .944          \\
\multicolumn{1}{l|}{5 MTL-DistilBERT} & 94.2\%          & .914          & \multicolumn{1}{l|}{.980}          & 90.4\%          & .902          & \multicolumn{1}{l|}{.964}          & 90.6\%          & .852          & \textbf{.951} \\
\bottomrule
\end{tabular}
\label{table:results}
\end{table*}

\section{Experiments And Results}

In this section, we first introduce training details and evaluation metrics and then show experimental results.

\subsection{Training}

\textbf{Train/Test split:} We find by experiments that increasing training size does not help the classifier when the number of training samples is over 5000. Therefore, 5000/2053/5000 data samples are used for training/validation/testing.

\vspace{-3pt}

\textbf{Loss Functions:} For BERT and DistilBERT based Single-Task Learning (STL) models, the cross-entropy loss is used. For BERT and DistilBERT base Multi-Task Learning (MTL) models, the cross-entropy loss is used for each task. The total loss will be the sum of the cross-entropy loss of each task.

\vspace{-3pt}

\textbf{Cost-Sensitive method:} As mentioned in Section 3.3, the dataset is mildly imbalanced (minority class > 20\%). Thus, a cost-sensitive method is used in this study for alleviating the problem of class imbalance and improving performance, by weighting the cross-entropy loss function during training based on the frequency of each class in the training set.

\vspace{-3pt}

\textbf{Hyperparameters:} As we mentioned in Section 4.2 and Section 4.3, we use HuggingFace pre-trained BERT and DistilBERT to initialize models. The hidden size for BERT and DistilBERT is 768. We then fine-tune the BERT and DistilBERT based single-task learning and multi-task learning models with a batch size of 32, max sequence length 100, learning rate 2e-5/3e-5/5e-5, epochs of 2/3, dropout rate 0.1, and Adam optimizer with $\beta_1$=0.9 and $\beta_2$=0.99.

\vspace{-3pt}

\subsection{Evaluation Metrics}
We use accuracy, macro-F1 score (average for each class of each label instead of each label), and AUC (Area Under ROC Curve) to evaluate models. Since the dataset is merely mildly imbalanced, accuracy can still be a useful metric. The Macro-F1 instead of F1-score for the positive class is used, since both positive class and negative class for each label are important for our task. For this study, we mainly use accuracy and macro-F1 to compare different models.

\subsection{Results}

Table \ref{table:results} shows the performance of all models when training with a different number of training samples (1K, 3K, and 5K). The first column indicates the models (GloVe, BERT, DistilBERT) and training settings (single-task learning (STL), multi-task learning (MTL)).

\textbf{RQ1 Does BERT outperform previous methods?} \\
We first implemented a baseline single-task learning model by leveraging pre-trained GloVe (Global Vectors for Word Representation)\footnote{https://nlp.stanford.edu/projects/glove/} \cite{pennington2014glove} word embeddings. We added a BatchNormalization layer on top of GloVe, and the aggregate representation of the input sequence for classification was obtained by AveragePooling the output of the BatchNormalization layer. A dense layer was added on the top for performing classification.

We compared GloVe and BERT for every single task. As shown in Table \ref{table:results}, the results clearly showed that a BERT-based STL model yields substantial improvements over the previous GloVe-based method. The STL-BERT model trained with 1000 data samples outperformed the STL-GloVe model trained with 5000 data samples on all tasks. This suggests that the need for labeled data could be significantly reduced by leveraging a pre-trained language model BERT.

\textbf{RQ2 How does multi-task learning perform?} \\
By comparing MTL-BERT with STL-BERT and MTL-Distil- BERT with STL-DistilBERT when trained with a different number of training samples, we found that jointly learning related tasks improves the performance of the suggestion-detection task and the positive-tone detection task, especially when we have limited training samples (i.e., when training with 1K and 3K data samples). This suggests that MTL can increase data efficiency. However, for the problem-detection task, there is no significant difference between the performance of the STL and MTL settings. 

Additionally, MTL can considerably reduce the model size. As shown in Table \ref{table:parameters}, three BERT-based STL models would have more than 328M parameters, and this number would be 199M for the DistilBert-based models. However, if we employ the MTL models for evaluating peer-review comments, the number of parameters would be reduced to 109M and 66M, respectively. This demonstrates that using MTL to evaluate reviews can save considerable memory resources and reduce the response time of peer-review platforms.

\begin{table}[]
\caption{The \# of parameters for each setting}
\centering
\begin{tabular}{ll}
\toprule
Setting            & \# of parameters \\
\midrule
STL-BERT * 3       & 328M             \\
STL-DistilBERT * 3 & 199M             \\
MTL-BERT           & 109M             \\
MTL-DistilBERT     & 66M              \\
\bottomrule    
\end{tabular}
\label{table:parameters}
\vspace{-10pt}
\end{table}

\textbf{RQ3 How does DistilBERT perform?}\\
By comparing DistilBERT and BERT on both STL and MTL settings, we found that BERT-based models slightly outperformed DistilBERT-based models. This result implied a trade-off between performance and model size when selecting the model to be deployed on peer-review platforms. If we focus on high accuracy instead of memory resource usage and response time of the platforms, the MTL-BERT model is the choice. Otherwise, the MTL-DistilBERT should be deployed.

\section{Conclusions}

In this study, we implemented single-task and multi-task models for evaluating peer-review comments based on the state-of-the-art language representation models BERT and DistilBERT. Overall, the results showed that BERT-based STL models yield significant improvements over the previous GloVe-based method on tasks of detecting a single feature. Jointly learning different tasks simultaneously further improves performance and saves considerable memory usage and response time for peer-review platforms. The MTL-BERT model should be deployed on peer-review platforms, if our focus is on high accuracy instead of memory resource usage and response time of the platforms. Otherwise, the MTL-DistilBERT model is preferred.

There are three limitations to this study. Firstly, we employed three features of high-quality peer reviews to evaluate a peer-review comment. However, it is still unclear how MTL will perform if we learn more tasks simultaneously. Secondly, we mainly focused on a hard-parameter sharing approach for constructing MTL models. However, some studies have found that the soft-parameter sharing approach might be a more effective method for constructing multi-task learning models. Thirdly, the performance of the model has not been evaluated in actual classes. We intend to deploy the model on the peer-review platform and evaluate the model extrinsically in real-world circumstances. 

These preliminary results serve as a basis for our ongoing work, in which we are building a more complex all-in-one model for comprehensively and automatically evaluating the quality of peer review comments to improve peer assessment. In the future, we will attempt to evaluate peer reviews based on more predetermined features and use fine-grained labels (e.g., instead of evaluating whether a peer-review comment contains suggestions, we will evaluate how many suggestions are contained in a review comment).

% \end{document}  % This is where a 'short' article might terminate

%ACKNOWLEDGMENTS are optional
%\section{Acknowledgments}
%This section is optional; it is a location for you
%to acknowledge grants, funding, editing assistance and
%what have you.  In the present case, for example, the
%authors would like to thank Gerald Murray of ACM for
%his help in codifying this \textit{Author's Guide}
%and the \textbf{.cls} and \textbf{.tex} files that it describes.

\bibliographystyle{abbrv}
\bibliography{sigproc}  % sigproc.bib is the name of the Bibliography in this case

\begin{thebibliography}{10}

\bibitem{brutus2013can}
S.~Brutus, M.~B. Donia, and S.~Ronen.
\newblock Can business students learn to evaluate better? evidence from
  repeated exposure to a peer-evaluation system.
\newblock {\em Academy of Management Learning \& Education}, 12(1):18--31,
  2013.

\bibitem{caligiuri2013editors}
P.~Caligiuri and D.~C. Thomas.
\newblock From the editors: How to write a high-quality review, 2013.

\bibitem{caruana1997multitask}
R.~Caruana.
\newblock Multitask learning.
\newblock {\em Machine learning}, 28(1):41--75, 1997.

\bibitem{cho2008machine}
K.~Cho.
\newblock Machine classification of peer comments in physics.
\newblock In {\em Educational Data Mining 2008}, 2008.

\bibitem{crawshaw2020multi}
M.~Crawshaw.
\newblock Multi-task learning with deep neural networks: A survey.
\newblock {\em arXiv preprint arXiv:2009.09796}, 2020.

\bibitem{devlin2018bert}
J.~Devlin, M.-W. Chang, K.~Lee, and K.~Toutanova.
\newblock Bert: Pre-training of deep bidirectional transformers for language
  understanding.
\newblock {\em arXiv preprint arXiv:1810.04805}, 2018.

\bibitem{double2020impact}
K.~S. Double, J.~A. McGrane, and T.~N. Hopfenbeck.
\newblock The impact of peer assessment on academic performance: A
  meta-analysis of control group studies, 2020.

\bibitem{fromm2020argument}
M.~Fromm, E.~Faerman, M.~Berrendorf, S.~Bhargava, R.~Qi, Y.~Zhang, L.~Dennert,
  S.~Selle, Y.~Mao, and T.~Seidl.
\newblock Argument mining driven analysis of peer-reviews.
\newblock {\em arXiv preprint arXiv:2012.07743}, 2020.

\bibitem{gehringer2007reusable}
E.~Gehringer, L.~Ehresman, S.~G. Conger, and P.~Wagle.
\newblock Reusable learning objects through peer review: The expertiza
  approach.
\newblock {\em Innovate: Journal of Online Education}, 3(5):4, 2007.

\bibitem{gupta2020compression}
M.~Gupta, V.~Varma, S.~Damani, and K.~N. Narahari.
\newblock Compression of deep learning models for nlp.
\newblock In {\em Proceedings of the 29th ACM International Conference on
  Information \& Knowledge Management}, pages 3507--3508, 2020.

\bibitem{li2020does}
H.~Li, Y.~Xiong, C.~V. Hunter, X.~Guo, and R.~Tywoniw.
\newblock Does peer assessment promote student learning? a meta-analysis.
\newblock {\em Assessment \& Evaluation in Higher Education}, 45(2):193--211,
  2020.

\bibitem{mchugh2012interrater}
M.~L. McHugh.
\newblock Interrater reliability: the kappa statistic.
\newblock {\em Biochemia medica}, 22(3):276--282, 2012.

\bibitem{negi2017suggestion}
S.~Negi and P.~Buitelaar.
\newblock Suggestion mining from opinionated text.
\newblock {\em Sentiment Analysis in Social Networks}, pages 129--139, 2017.

\bibitem{nelson2009nature}
M.~M. Nelson and C.~D. Schunn.
\newblock The nature of feedback: How different types of peer feedback affect
  writing performance.
\newblock {\em Instructional Science}, 37(4):375--401, 2009.

\bibitem{nguyen2014classroom}
H.~Nguyen, W.~Xiong, and D.~Litman.
\newblock Classroom evaluation of a scaffolding intervention for improving peer
  review localization.
\newblock In {\em International Conference on Intelligent Tutoring Systems},
  pages 272--282. Springer, 2014.

\bibitem{nguyen2016instant}
H.~Nguyen, W.~Xiong, and D.~Litman.
\newblock Instant feedback for increasing the presence of solutions in peer
  reviews.
\newblock In {\em Proceedings of the 2016 Conference of the North American
  Chapter of the Association for Computational Linguistics: Demonstrations},
  pages 6--10, 2016.

\bibitem{omar2018use}
D.~Omar, M.~Shahrill, and M.~Zuraifah~Sajali.
\newblock The use of peer assessment to improve students’ learning of
  geometry.
\newblock {\em European Journal of Social Science Education and Research},
  5(2):187--206, 2018.

\bibitem{pennington2014glove}
J.~Pennington, R.~Socher, and C.~D. Manning.
\newblock Glove: Global vectors for word representation.
\newblock In {\em Proceedings of the 2014 conference on empirical methods in
  natural language processing (EMNLP)}, pages 1532--1543, 2014.

\bibitem{ramachandran2013automated}
L.~Ramachandran et~al.
\newblock Automated assessment of reviews.
\newblock 2013.

\bibitem{sadler2006impact}
P.~M. Sadler and E.~Good.
\newblock The impact of self-and peer-grading on student learning.
\newblock {\em Educational assessment}, 11(1):1--31, 2006.

\bibitem{sanh2019distilbert}
V.~Sanh, L.~Debut, J.~Chaumond, and T.~Wolf.
\newblock Distilbert, a distilled version of bert: smaller, faster, cheaper and
  lighter.
\newblock {\em arXiv preprint arXiv:1910.01108}, 2019.

\bibitem{suen2014peer}
H.~K. Suen.
\newblock Peer assessment for massive open online courses (moocs).
\newblock {\em International Review of Research in Open and Distributed
  Learning}, 15(3):312--327, 2014.

\bibitem{sun2019fine}
C.~Sun, X.~Qiu, Y.~Xu, and X.~Huang.
\newblock How to fine-tune bert for text classification?
\newblock {\em arXiv preprint arXiv:1905.05583}, 2019.

\bibitem{topping1998peer}
K.~Topping.
\newblock Peer assessment between students in colleges and universities.
\newblock {\em Review of educational Research}, 68(3):249--276, 1998.

\bibitem{van2010effective}
M.~Van~Zundert, D.~Sluijsmans, and J.~Van~Merri{\"e}nboer.
\newblock Effective peer assessment processes: Research findings and future
  directions.
\newblock {\em Learning and instruction}, 20(4):270--279, 2010.

\bibitem{vaswani2017attention}
A.~Vaswani, N.~Shazeer, N.~Parmar, J.~Uszkoreit, L.~Jones, A.~N. Gomez,
  L.~Kaiser, and I.~Polosukhin.
\newblock Attention is all you need.
\newblock {\em arXiv preprint arXiv:1706.03762}, 2017.

\bibitem{violato2006self}
C.~Violato and J.~Lockyer.
\newblock Self and peer assessment of pediatricians, psychiatrists and medicine
  specialists: implications for self-directed learning.
\newblock {\em Advances in Health Sciences Education}, 11(3):235--244, 2006.

\bibitem{wang2012assessment}
Y.~Wang, H.~Li, Y.~Feng, Y.~Jiang, and Y.~Liu.
\newblock Assessment of programming language learning based on peer code review
  model: Implementation and experience report.
\newblock {\em Computers \& Education}, 59(2):412--422, 2012.

\bibitem{wu2016google}
Y.~Wu, M.~Schuster, Z.~Chen, Q.~V. Le, M.~Norouzi, W.~Macherey, M.~Krikun,
  Y.~Cao, Q.~Gao, K.~Macherey, et~al.
\newblock Google's neural machine translation system: Bridging the gap between
  human and machine translation.
\newblock {\em arXiv preprint arXiv:1609.08144}, 2016.

\bibitem{xiaoproblem}
Y.~Xiao, G.~Zingle, Q.~Jia, S.~Akbar, Y.~Song, M.~Dong, L.~Qi, and
  E.~Gehringer.
\newblock Problem detection in peer assessments between subjects by effective
  transfer learning and active learning.

\bibitem{xiao2020detecting}
Y.~Xiao, G.~Zingle, Q.~Jia, H.~R. Shah, Y.~Zhang, T.~Li, M.~Karovaliya,
  W.~Zhao, Y.~Song, J.~Ji, et~al.
\newblock Detecting problem statements in peer assessments.
\newblock {\em arXiv preprint arXiv:2006.04532}, 2020.

\bibitem{xiong2010identifying}
W.~Xiong and D.~Litman.
\newblock Identifying problem localization in peer-review feedback.
\newblock In {\em International Conference on Intelligent Tutoring Systems},
  pages 429--431. Springer, 2010.

\bibitem{xiong2011automatically}
W.~Xiong and D.~Litman.
\newblock Automatically predicting peer-review helpfulness.
\newblock In {\em Proceedings of the 49th Annual Meeting of the Association for
  Computational Linguistics: Human Language Technologies}, pages 502--507,
  2011.

\bibitem{xiong2010assessing}
W.~Xiong, D.~Litman, and C.~Schunn.
\newblock Assessing reviewers' performance based on mining problem localization
  in peer-review data.
\newblock In {\em Educational Data Mining 2010-3rd International Conference on
  Educational Data Mining}, pages 211--220, 2010.

\bibitem{zhang2017survey}
Y.~Zhang and Q.~Yang.
\newblock A survey on multi-task learning.
\newblock {\em arXiv preprint arXiv:1707.08114}, 2017.

\bibitem{zhang2021survey}
Y.~Zhang and Q.~Yang.
\newblock A survey on multi-task learning.
\newblock {\em IEEE Transactions on Knowledge and Data Engineering}, 2021.

\bibitem{zingle2019detecting}
G.~Zingle, B.~Radhakrishnan, Y.~Xiao, E.~Gehringer, Z.~Xiao, F.~Pramudianto,
  G.~Khurana, and A.~Arnav.
\newblock Detecting suggestions in peer assessments.
\newblock {\em International Educational Data Mining Society}, 2019.

\end{thebibliography}
% ACM needs 'a single self-contained file'!
%\balancecolumns

\end{document}